\documentclass[letterpaper, 10 pt, conference]{ieeeconf}  

\IEEEoverridecommandlockouts                              

\usepackage{cite}
\usepackage{amsmath,amssymb,amsfonts}
\usepackage{algorithmic}
\usepackage{graphicx}
\usepackage{textcomp}
\usepackage{xcolor}
\usepackage{todonotes}
\usepackage{cite}
\usepackage{tabularx}
\usepackage{array}
\usepackage{amsmath}
\usepackage{multirow}
\usepackage{caption}
\usepackage{subcaption}

\def\BibTeX{{\rm B\kern-.05em{\sc i\kern-.025em b}\kern-.08em
    T\kern-.1667em\lower.7ex\hbox{E}\kern-.125emX}}
\begin{document}

\title{\LARGE \bf See as a Bee: UV Sensor for Aerial Strawberry Crop Monitoring
}

\author{Megan Heath$^{2}$, Ali Imran$^{1}$ and David St-Onge$^{1}$
\thanks{*We thank NSERC CREATE program UTILI for their financial support.}
\thanks{$^{1}$Ali Imran and David St-Onge are with the Lab INIT Robots, Department of Mechanical Engineering,
        Ecole de technologie supérieure, Canada
        {\tt\small name.surname@etsmtl.ca}}%
\thanks{$^{2}$Megan Heath is with the Department of Environmental Engineering, Ecole de technologie supérieure,
        Canada
        {\tt\small megan.heath.1@ens.etsmtl.ca}}%
}

\maketitle
\thispagestyle{empty}
\pagestyle{empty}

\begin{abstract}
Precision agriculture aims to use technological tools for the agro-food sector to increase productivity, cut labor costs, and reduce the use of resources. This work takes inspiration from bees’ vision to design a remote sensing system tailored to incorporate UV-reflectance into a flower detector. We demonstrate how this approach can provide feature-rich images for deep learning strawberry flower detection and we apply it to a scalable, yet cost effective aerial monitoring robotic system in the field. We also compare the performance of our  UV-G-B image detector with a  similar work that utilizes RGB images.

\end{abstract}

\section{Introduction}
Agriculture is known to be one of most active sectors for innovation. 
Now more than ever, with the changing dynamics of the world due to climate change, urbanization and the increased human population, the agriculture sector needs to adapt and increase adoption of automated systems for crop management. This (r)evolution is meant to happen; according to Tilan et al.\cite{tilman2011global}, the need for the three basic crops (wheat, maize and rice) will increase by 110\% by 2050.

Several technologies support the required innovations, namely mobile robotic systems such as unmanned aerial vehicles (UAVs). As the accessibility to these platforms increases, UAVs are quickly becoming the tool of choice across agricultural \cite{tsouros2019review} and non-agricultural \cite{shakhatreh2019unmanned} sectors for data gathering.
Thanks to the availability of various remote sensors enabling capture, process and analysis of airborne data to provide farmers with accurate information about their crops and help them make more informed decisions. This also directs needs-based application of water, nutrients, and chemicals. Such is the basis of precision agriculture. In parallel, the industry underwent major development in the domain of robotized crops manipulation, such as harvesters and even pollinators. The latter is motivated by the near-extinction of several species of bees in some parts of the world. According to Aurell et al. \cite{aurellunited}, the population of honey bees has decreased of 23.8\% in the year 2021-2022. Moreover, some of the modern agricultural practices, often beneficial to the industry and society, provide an unsuitable environment for natural pollinators (bees), such as greenhouses and poly-tunnels. For instance, the growing popularity of urban agriculture and vertical farms \cite{goldstein2018green}, calls for creative innovation for sustainable pollination of the crops.

The development of digital cameras, using CCD and CMOS sensors, more powerful computers, and object detection algorithms have been key to precision agriculture. Today, neural network algorithms combined with RGB or hyperspectral camera data has given rise to vegetation indices that can measure soil and plant health, crop growth, and nutrient requirements \cite{kattenborn2021review}. It is also showing great potential for weed and disease detection, input requirements, and crop yield estimation \cite{tsouros2019review}. 

In order to support artificial pollination, the robotic system first need to detect and localise the target flowers. This feature also provides data for crop counting or flower counting for yield estimation. Manual stand counts are labor intensive and prone to human error. By using a UAV mounted camera and AI software, automated, non-subjective yield estimation can be achieved at a fraction of the cost and effort. Most of the existing remote sensing systems are based on human vision (RGB); mimicking farmers visually observing their fields. However, flowering plants have co-evolved for millennia to interact with insects whose vision spans the UV-G-B range \cite{briscoe2001evolution}. Pollinating insects such a honey bees (Apis sp.) can distinguish crop species and cultivars from one another based on floral patterning undetectable to human vision \cite{briscoe2001evolution}, against a complex background, and while airborne. This is due to the greater contrast seen in the UV-G-B spectrum than the RGB. Our hypothesis is that the UV-G-B spectrum would be better suited to UAV platforms for flower detection as it would mimic pollinator vision and detect intended plant cues for aerial pollinators.

This paper presents a biologically inspired  
 UV-G-B camera (Sec. ~\ref{sec:sensor}), which mimics a natural pollinators' vision, for detecting crop flowers using state-of-the-art object detection algorithms tuned to the task (Sec.~\ref{sec:algo}. We then validate the technology with a complete deployment onboard a UAV in a strawberry field (Sec.~\ref{sec:exp}), and show that our detector yields better results compared to another similar deployment which used RGB images.

\section{Related Work}\label{sec:litrev}

From huge phenotyping platforms such as Field Scanalyzer \cite{virlet2016field}, to unmanned ground and aerial vehicles (UGVs and UAVs), many platforms are currently used for gathering field data. There are already several options of ground vehicles available for sensing and phenotyping platforms \cite{deery2014proximal,williams2020improvements}. Nevertheless, recent years have seen a tendency towards the use of UAV for precision agriculture \cite{mulla2013twenty,kim2019unmanned}. Depending on the application, some aerial platforms offer a better set of features than others. Fixed wing platforms have greater flight time and more payload capacity, however, it is difficult to get higher quality data due to limited hovering capabilities. Blimps offer the advantage of simpler operations but it becomes difficult to operate in harsh outdoor conditions. Rotor copters have better hovering capabilities and thus offer a better chance to capture higher quality imagery. However, these platforms have limited flight times \cite{sankaran2015low}. It also presents a very comprehensive review of the advantages and disadvantages of different types of platforms for applications in the domain of agriculture. As for artificial pollinators, UGV based robotic systems have been already proposed for Kiwi fruit pollination \cite{williams2020autonomous,li2022identification} and used in poly-tunnels \cite{le2020low,ko2014autonomous}.


The most frequently used remote sensors for precision agriculture applications are visible light sensors (RGB). These light weight, inexpensive sensors replicate the human vision range (400-700nm) when capturing images and they benefit from the largest literature on object detection algorithms. Detecting flowers and fruits using these sensors have led to different types of algorithms, such as flower contour detection through color and edge detection \cite{hong2012automatic}, and spectral spatial methods for e.g. detection of tomatoes \cite{senthilnath2016detection}. However, their main disadvantage is the inability to analyze any parameter outside the visual spectrum \cite{tsouros2019review}. Other spectral bands offer valuable data about the crops, such as the state of chlorophyll, which reflects near infrared radiation (NIR). Naturally, when a plant becomes diseased or stressed, the amount of chlorophyll reduces - resulting in a an overall changed spectrum within this range. Studies have shown that UAV based NIR imagery can be used to accurately detect disease and water stress in crops \cite{antolinez2022identification,zhou2021assessment}. 

As such, the use of multispectral 
cameras in agriculture is growing in popularity. With a Visible Near-Infrared (Vis-NIR) camera, a robotic system can detect powdery mildew in asymptomatic squash plant with 89\% accuracy under field conditions \cite{abdulridha2020detecting}. Similar sensor can also serve to study water stress in apple orchards showing a strong correlation (R² = 0,9975) between temperature measured on ground targets and estimates made from aerial images \cite{gomez2016field}. Stumph et al. \cite{stumph2019detecting} used UV LED lights on a UAV with an RGB camera to induce fluorescence in tree dwelling insects and they achieved detection precision as high as 80\%. Thus the complete plant reflection spectra, acquired using multispectral and hyperspectral cameras, is key to monitoring several important plant health and growth attributes.

\begin{figure}[htp]
    \centering
    \includegraphics[width=\columnwidth]{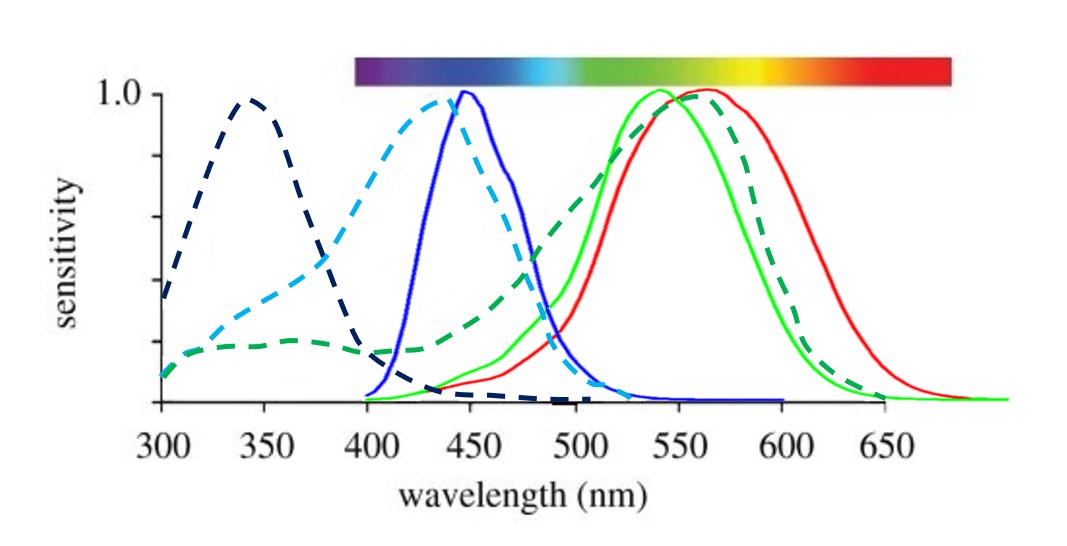}
    \caption{Spectral Sensitivity of Western Honey Bee compared to Human vision range. Bee Uv, B, and G  are represented by the dotted black, blue and green lines respectively.Human sensitivity for Red, Green, and Blue correspond to the solid line colors \cite{dyer2015seeing, coliban2020linear}.}
    \label{fig:bee1}
\end{figure}


However, these cameras are expensive, heavy and their wide spectrum is mostly underused for specific crop monitoring tasks. Taking inspiration from biological pollinating agents such as honey bees, a spectrum of near-UV, Green, and blue light is used to perceive and identify flowers \cite{briscoe2001evolution}. The lack of development of such sensors may come form the lack of studies exploring the use of UV floral reflectance of plants; let alone crops. Spectral reflectance studies have reported down to 300nm but most species in reflectance databases, are native species not agricultural crops \cite{arnold2008fred}. Furthermore, UV cameras and lenses in these studies were bulky and very expensive; leading to even fewer field studies \cite{stuart2019hyperspectral}.

This project aims at developing an efficient, biologically inspired flower detection system, which leverages the UV-G-B spectrum using a powerful yet cost effective drone setup.

\section{UV-G-B sensor design}\label{sec:sensor}
The design of our robotic system was tested on a representative of the Rosacea family of crops; the strawberry. Canada produces 24,185 metric tons of strawberries annually; making it the 5th most valued fruiting crop for Canada in 2020 \cite{canadastrawberry}. Unlike Rosacea orchard species (e.g., apple,) some commercial strawberry cultivars are everbearing; flowering throughout the growing season. This, along with its compact habit and flowers held on stocks above the foliage, made it an ideal crop to use for aerial imagery in this study. Strawberry flowers are mainly pollinated by various bee species; primarily Apis mellifera, or western honey bees, in agricultural settings  \cite{James123}. Like humans, bees perceive in 3 light ranges. While humans see from 400-700nm, from blue to red (RGB), honey bees see from 300-650nm, near-UV to green \cite{briscoe2001evolution}. Both sensitivity curves are plot in fig. \ref{fig:bee1}. Most digital cameras are designed to capture images in the human visual spectrum, e.g. RGB. This is achieved by coating the sensor with a Bayer filter which is a pattern of red, green and blue filters across the photodiode matrix \cite{palum2001image}.

In order to create a camera which can capture images similar to pollinator vision, the light reaching the sensor must be restricted to the UVGB range. Also, to acquire usable data, the aerial vision system have to provide a ground sampling distance (GSD) comparable to existing flower detection aerial imagery; ideally in the range of 0.7 to 3 cm/pixel\cite{vanbrabant2020pear, chen2019strawberry, hunt2018good}. GSD is computed with: 
\[\ GSD = \lambda \frac{H}{c}\]
where $H$ is the flight altitude (m), $c$, the Focal length (mm), and $\lambda$, the camera sensor pixel size. 

\subsection{UV-G-B Camera Design}
We designed light weight, affordable, and drone mountable camera after the vision range of most insect pollinators: 300-650nm \cite{briscoe2001evolution}.  

The sensor is based on a MaxMax XNite USB8M-M \cite{maxmaxcam}, which provides a video capture resolution of 3264x2448. Considering a flight height of 3m, we can expect a GSD = 1.16cm/pixel. These cameras retail for ~500 USD and are highly customizable to adapt to the specific needs of the application. The X-Nite camera is most comparable to a GoPro \cite{goprocamera} in terms of size and weight. See Table \ref{table:sensorcomparison} for detailed comparison.

\begin{table}[h]
\caption{Our camera model contrasted with a comparable camera on the market}
\begin{tabular}{ | m{3em} | m{1cm}| m{1cm} | m{1cm} | m{1cm} | m{1cm} | } 
  \hline
  Camera& Body Weight (g) & Lens Weight (g) & Body Dim. (LxWxH) cm & Cable (15cm) Weight (g) & Total \\ 
  \hline
  X-Nite & 83 & ~4 & 4x 2.2 x 4 & ~7 & 87-91g \\ 
  \hline
  GoPro HERO 9 & 158 & na & 5.5 x 7.1 x 3.3 & na & 158g\\
  \hline
\end{tabular}
\label{table:sensorcomparison}
\end{table}

\begin{figure}[htp]
    \centering
\includegraphics[width=0.9\columnwidth]{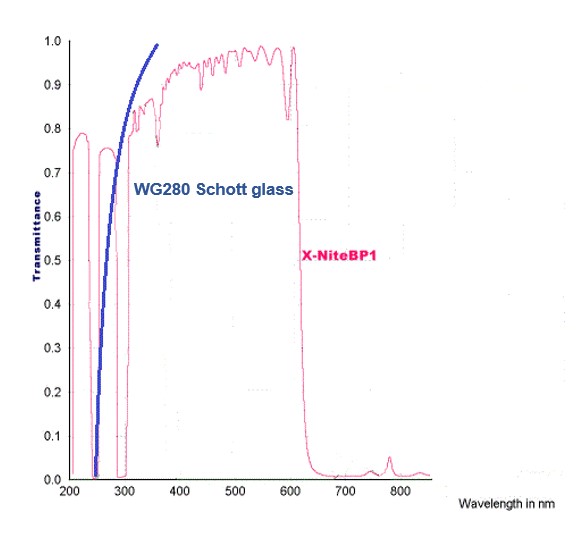}
    \caption{Transmission spectrograph for Schott glass (blue) and for XNite BP1 filter (pink). Combination allows upwards of 80 \% transmission  in the 300-650nm range.  }
    \label{fig:schott}
\end{figure}

Unlike the GoPro, the X-Nite has a USB cable which allows the camera to be connected directly to a computer for image capture and continuous power draw.
In order for the camera range to extend to the UV range (300-400nm), the optical window, or ICF, was replaced with Schott WG280 glass which transmits light down to 280nm \cite{graph1}. The glass of the lense was also replaced with Schott WG280 glass and had no laminated filter installed. To restrict the upper range of the spectrum, the X-NiteBP1 filter from Max Max was chosen. It has 80-100\% transmission between 300-650nm; ideal for imitating insect vision in a monochrome image. Figure \ref{fig:schott} show the effect of the lens and filter to the sensor's spectrum sensitivity.

The camera's CMOS sensor had its Bayer filter and micro lens removed; increasing transmission down to 300nm. Removal of the Bayer filter creates a single-channel monochrome sensor. Inherently, monochrome sensors can achieve higher resolution, faster processing times or frame rates, and be stored as smaller files than their RGB counterparts. It also means smaller processing time as colored images have to be processed over three dimensions. The Bayer filter itself can reduce the optical resolution of the system \cite{sensor1}. Since our use case requires the removal of the red filter, the Bayer filter was completely removed to optimize the camera’s performance. 


\subsection{Camera Characterization}
\begin{figure}[htp]
    \centering
    \includegraphics[width=\columnwidth]{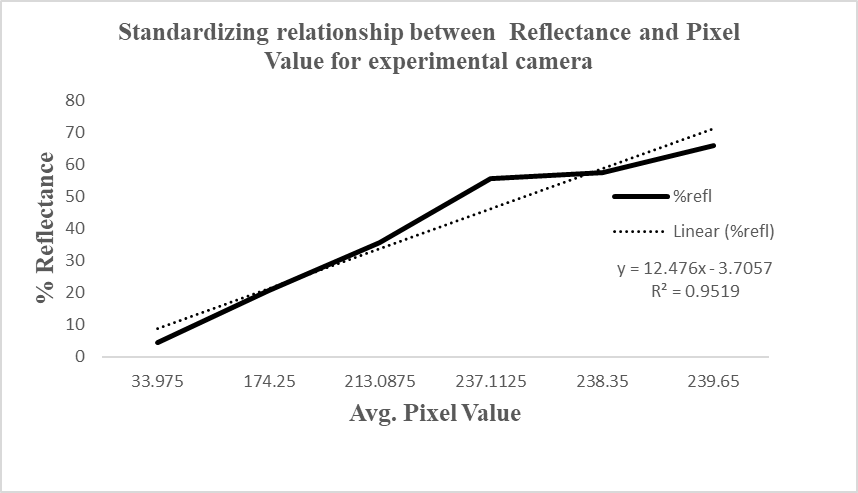}
    \caption{Relation between \% Reflectance of standard samples and pixel value produced by our camera sensor.}
    \label{fig:pix}
\end{figure}

\begin{figure}[htp]
    \centering
    \includegraphics[width=\columnwidth]{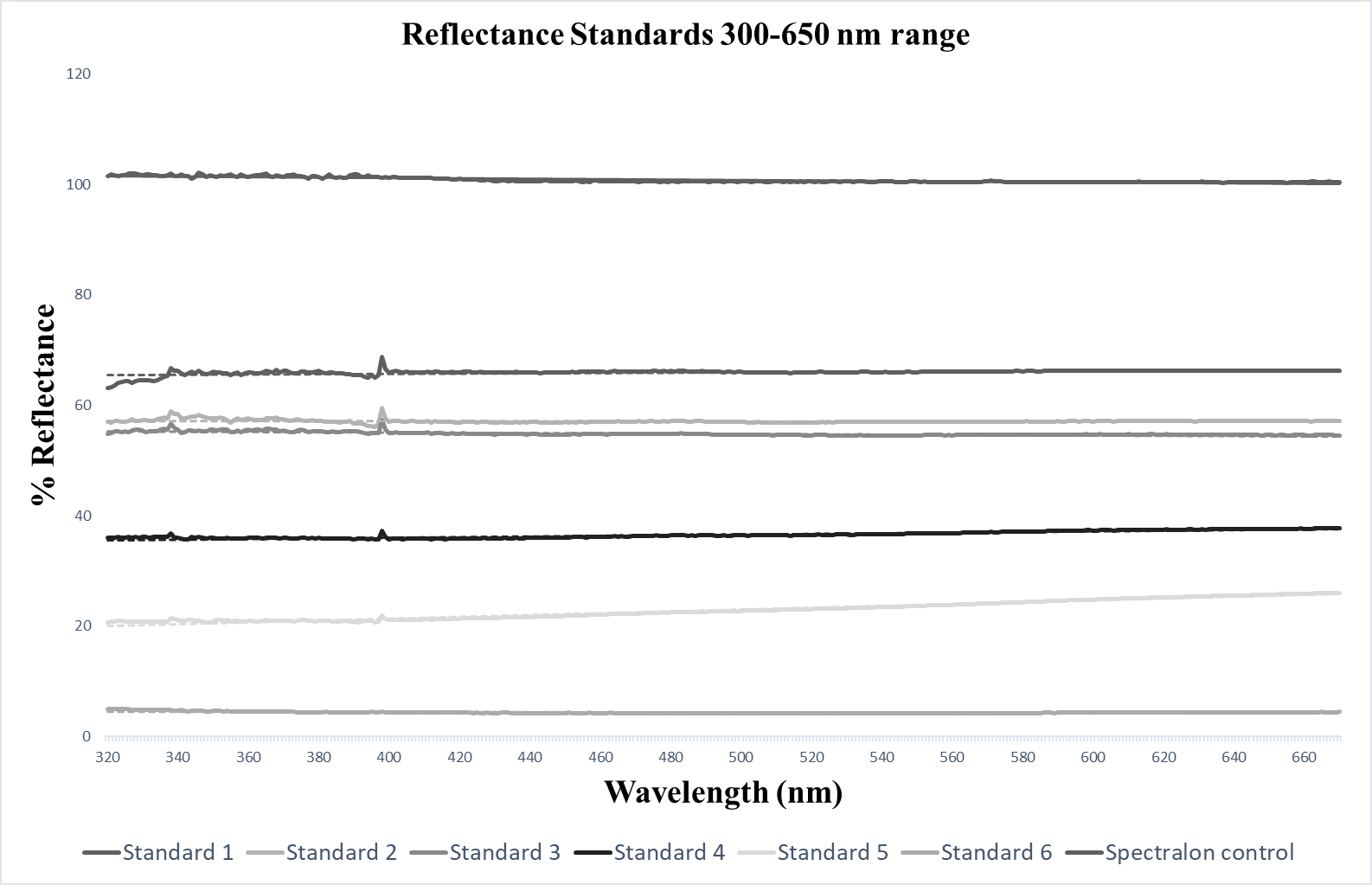}
    \caption{Spectrogram of reflectance standard samples with spectralon as a control}
    \label{fig:spec}
\end{figure}

As commercially available reflectance standards reflect UV (320-400 nm) poorly, we created an appropriate reflectance standard following the work of \cite{dyer2004calibrated}. Five standards were made from varying proportions of medical-grade MgO, Plaster of Paris, and activated carbon \cite{dyer2004calibrated}; we also added a sixth standard consisting of a black UV-absorbing plastic. Each standard’s (1-6) spectral reflectance (\% R) in the range 200nm - 700nm was measured with a Perkin Elmer’s Lambda 850 UV-VIS spectrophotometer. Figure \ref{fig:spec} shows the resulting spectrogram. The intercept for each curve denotes the consistent \% R each standard will emit across varying lighting conditions.Images of the standards were then taken with our camera and an average pixel value for each sample was attained (avg. of 10 pixels). Fig. \ref{fig:pix} shows the strong linear relationship  \(R^2=0.9821\) between \% R and the pixel value our sensor produces.

\section{Learning to see flowers}\label{sec:algo}
No single or two step detection algorithms has been tested on UV-G-B flower images. All pre-trained algorithms mentioned in Sec.~\ref{sec:litrev} also only use RGB image datasets. It is therefore necessary to create a training dataset of UV-G-B images for our target specie and sensor. Validate can the be performed to determine which algorithm is the best candidate for flower detection in this context.

On competition data sets Convolutional Neural Networks (CNN) have proven themselves to be the best approach for target identification and classification due to their ability to extract increasingly complex visual features through its hierarchical structure \cite{zheng2021remote}. In regards to strawberry flower counting, Faster Region-based Convolutional Neural Networks (RCNN) have shown best results (86.1\%, 86.4\%, and 86\% accuracy respectively from \cite{lin2018detection,chen2019strawberry}, and \cite{zhou2020strawberry}). In an aerial application \cite{chen2019strawberry} recently employed Faster R-CNN for strawberry flower detection with a DJI phantom 4 pro and attained mAP= 0.772 \cite{chen2019strawberry}. Given their flight altitude of 3m and mixed training data set of strawberry cultivars this is the most recent comparable work to our current study. 

However, Faster RCNN often has a longer processing time \cite{mahendrakar2022performance} as compared to Yolo. Furthermore, Immaneni et al. \cite{immaneni2022real} tested YOLOV4 \cite{bochkovskiy2020yolov4} drone images from a strawberry field and achieved a better accuracy (91.95\% at 14.6 FPS). Related species such as Pear flowers, have also shown promising results with Yolo (mAP of 94\%) \cite{wang2022study}. 


\subsection{UV-G-B Strawberry Flower Dataset}

Since no image datasets of UVGB images are available, we create an original dataset using various strawberry cultivars. All datasets generated from this study are openly available on the Roboflow platform and can be accessed throught the IEEE dataport. F. x ananassa ‘Seascape’ and ‘Fort Laramie’ are commonly recommended commercial everbearing cultivars for central and east coast Canadian provinces \cite{strawberryvarieties}. Therefore, they were included along with F. x ananassa ‘Hecker’; which is an older everbearing cultivar which was popular with commercial growers for decades prior to ‘Seascape’ \cite{strawberryvarieties}. F. Vesca is a North American woodland native grown by breeders for its genetic attributes and specialty fruit growers for wine. All bare root plants were procured from a local Quebec nursery.

\begin{figure}[htp]
    \centering
    \includegraphics[width=1\columnwidth]{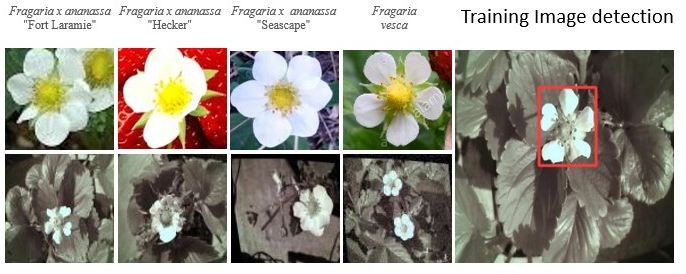}
    \caption{From left to right: strawberry cultivars 'Fort Laramie', 'Hecker', 'Seascape', and wild strawberry sp. vesca. Top row images are taken with RGB camera; bottom row images are with UVGB camera. The far right image shows Yolo detection of a strawberry flower on the training dataset. }
    \label{fig:allflowers}
\end{figure}

Images were captured under sunny conditions between 11am and 1pm within the 300-630nm. The camera was stationed 6cm from an open flower at a 90-degree overhead angle. Images were captured with 640x480 resolution (focal length= 3.6mm, GSD= 0.023 cm/pixel). An exterp of the dataset is shown in fig.~\ref{fig:allflowers}.Roboflow Inc. \cite{roboflowyolo} was used for data management and bounding box image labeling.
Images were Preprocessed for auto orientation and resized into a square shape, 416x416, for detection algorithm  compatibility. The following augmentations of the dataset were then applied: horizontal and vertical flip, rotate ±90-degrees, rotate ±15-degrees, sheer ±15-degrees vertical and horizontal, noise 5\%, and blur 5px. The resulting dataset consisted of 284 UV-G-B images.

We based our code implementation on publicly available script for both YoloV5 \cite{glenn_jocher_2020_4154370} and Faster R-CNN \cite{ren2015faster}. Pretrained weights for both algorithms were initially taken from MS COCO2017 \cite{lin2014microsoft}. We compared the algorithms using mAP@0.5, True positive (TP), False positive (FP), and False negative (FN) metrics. YoloV5s version was chosen due to the small training dataset in this study and ran for 1000 epochs. Faster R-CNN was run with inception V2 for 20000 steps.Training/validation/test split was 40\%/40\%/20\%.Table \ref{table:results416} shows the results for the trained algorithms. YoloV5s showed overall higher performance than Faster R-CNN and was therefore used for field testing of the systemn. 

\begin{table}[h]
\caption{Resulting detection from UVGB trained YOLOV5 and Faster R-CNN on training dataset at 416x416 resolution, and, on aerial images at 96x96 resolution }
\begin{tabular}{ | m{2.9em} | m{1cm}| m{0.6cm} | m{0.9cm} | m{0.8cm} | m{0.8cm} | m{0.8cm} | } 
  \hline
Training Dataset  & Detection Method & mAP @0.5 & Detection count & FP(Rate) & TP(Rate) & FN(Rate)\\
   \hline
   96x96 & YOLOv5 & 0.951 & 3260 & 2042 (62.2\%) & 1218 (37.1\%) & 25 (0.76\%) \\
   \hline
   96x96 & FR-CNN & 0.934 & 166 & 108(8.0\%) & 58(4.3\%) & 1185 (87.7\%) \\
   \hline
   416x416 & YOLOv5 & 0.978 & 77 & 0 (0\%) & 77 (97.4\%) & 2 (2.53\%) \\
  \hline
  416x416 & FR-CNN V2 & 0.912 & 144 & 69 (46.31\%) & 75 (50.34\%) & 5 (3.36\%) \\
\hline
\end{tabular}
\label{table:results416}
\end{table}

\section{Field deployment}\label{sec:exp}

To validate the usefulness of our solution, we deployed our sensor on a commercial UAV and flew over a local strawberry field.

\subsection{Aerial System implementation}
\begin{figure}[htp]
    \centering
    \includegraphics[width=\columnwidth]{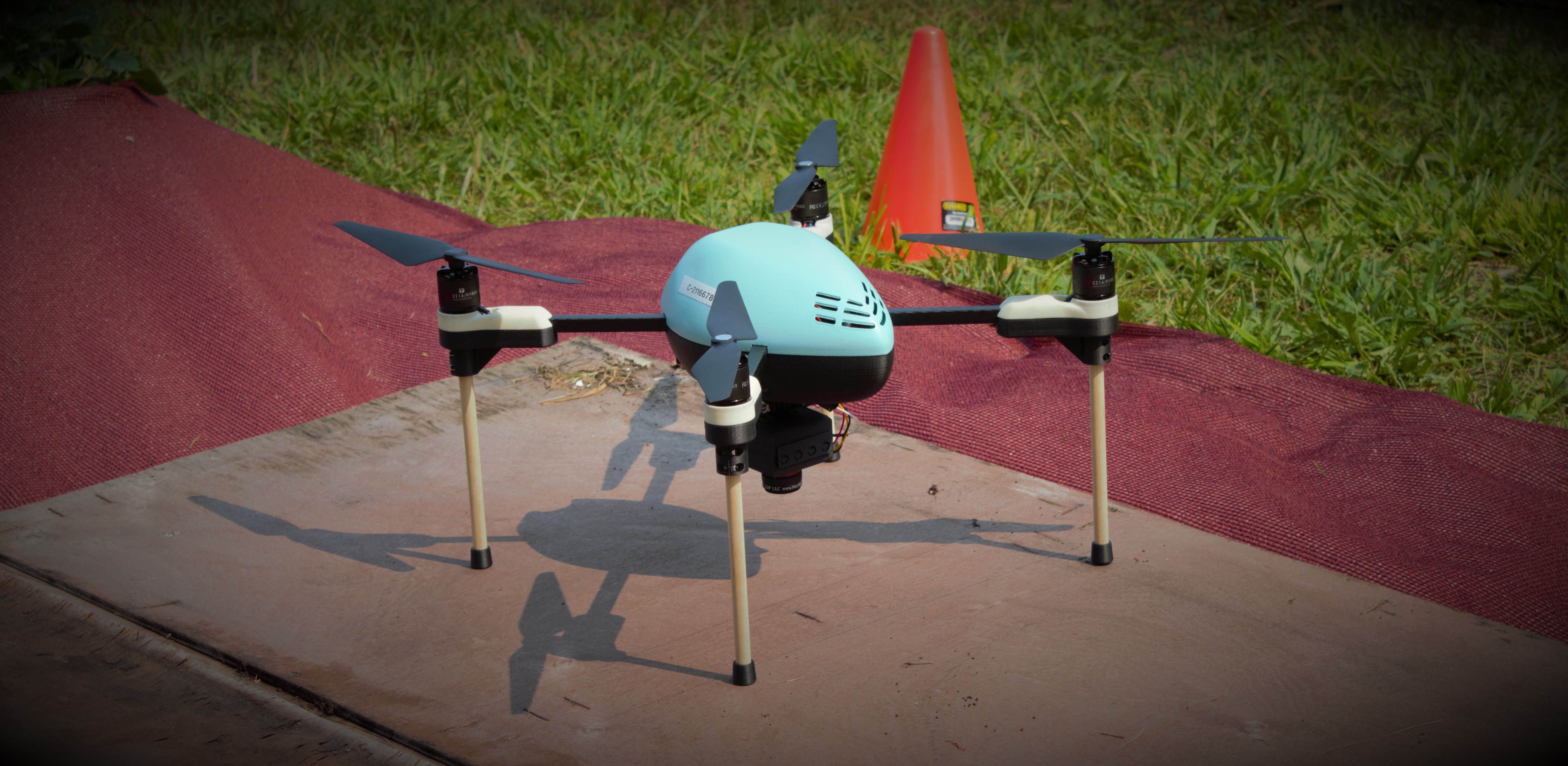}
    \caption{Our Spiri Mu prototype with the near-UV camera sensor pointing down and extended legs.}
    \label{fig:Spiri drone}
\end{figure}

We selected the UAV platform as to maximise the payload capacity while minimizing its cost and size. Spiri Mu \cite{spiriweb} stood out as the best option compared to other commercially available devices. Table \ref{table:drones comparison} details the comparison over the potential options we considered.
The Mu is powered by an Nvidia's TX2, powerful enough for heavy onboard image processing. To mount our camera and interface it to the onboard computer, we replaced the original underbelly of the Mu with an in-lab 3D printed attachment which could house the mounting for the camera. Moreover, the original landing gear of the drone was replaced by wooden dowels to make it taller in order to accommodate the additional sensor. Figure \ref{fig:Spiri drone} shows the modified drone in the field ready for take off.

The drone is connected to Qgroundcontrol (QGC) for calibrating onboard sensors and monitoring mission parameters during flight. The onboard system runs on Linux with ROS preinstalled which simplify the software integration of our sensors, and provide us with tools to record and transmit data efficiently. Our ROS camera driver is based on common ROS packages to fetch the USB camera feed and convert it to a ROS image topic.

\subsection{Field deployment setup}
Field flights were conducted at Pepinerie F. Fortier near Princeville, Qc. containing white-flowering ‘Seascape’ and ‘Albion’ cultivars. To keep consistent with the training dataset, video was captured under sunny conditions from 11am-1pm. An altitude of 3m reduced the effects of rotor downdraft on plants and produced a GSD=1.16cm/pix. Spanning 3 crop rows, 88 frames were captured and 2295 flowers were counted by a single reviewer. Frames were tiled and re-sized to 768x768 pixels to reduce processing time.

\subsection{Flower detection on UAV images}
 
Once aerial images were attained, the flower resolution was much lower than for the training dataset. YoloV5 and Faster R-CNN was therefore trained on the same image dataset as before but at a lower resolution, 96x96 pixels, to increase TP detection. Inferences were run with 0.51 confidence threshold on a Tesla P100-PCIE - 16 GB in Google colab \cite{Bisong2019}. Table \ref{table:results416} and table \ref{table:compareresutls} show the results of our trained YOLOV5 on the aerial images.





\begin{table}[h]
\caption{Cost and size comparison of similar UAV models on the market with the Spiri Mu \cite{spiriweb,mavric3specs,matrice300specs,phantom4specs} }
\begin{tabular}{ | m{4em} | m{2.4cm}| m{2cm} | m{1cm} | m{1.5cm} | } 
  \hline
  Drone & Dimensions LxWxH (mm) & Payload Capacity (g) & Base Cost (CAD) \\ 
  \hline
  Spiri Mu  & 170x170x51 & 1000 & 2000 \\ 
  \hline
   DJI M300  & 810×670×430 & 2700 & 12,722\\
  \hline
  DJI Mavic 3  & 347.5×283×107.7 & 727.4 & 2544 \\
  \hline
  DJI Phantom 4  & 289.5x289.5x196 & 800 & 2383 \\
  \hline
\end{tabular}
\label{table:drones comparison}
\end{table}



\begin{table}[h]
\caption{ Our UVGB camera detection results on aerial images compared to those from\cite{chen2019strawberry}}
\begin{tabular}{ | m{3em} | m{1.5cm}| m{1.5cm} | m{1.5cm} | m{1.2cm} |  } 
  \hline
UAV Model & Detection Method & Training Time (Hrs) & mAP@0.5 & FPS \\
  \hline
Spiri Mu & YOLOv5 & 0.3 & 0.951 & 0.0083 \\
  \hline
  Spiri Mu & FR-CNN Inc.V2 & 4.5 & 0.934 & 1.86 \\
  \hline
Phantom 4 Pro & FR-CNN with Resnet50 & 5.5 & 0.772 & 8.872\\
\hline
\end{tabular}
\label{table:compareresutls}
\end{table}

\subsection{Results}

\begin{figure}
     \centering
     \begin{subfigure}[b]{0.2\textwidth}
        \centering
         \includegraphics[width=\textwidth]{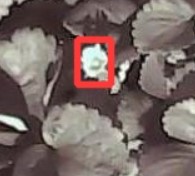}
         \caption{Flower}
         \label{fig:flower}
     \end{subfigure}
     \hfill
     \begin{subfigure}[b]{0.25\textwidth}
        \centering
         \includegraphics[width=\textwidth]{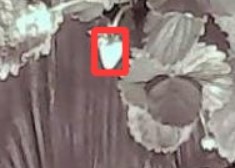}
         \caption{Fruit}
         \label{fig:strawberry}
     \end{subfigure}
        \caption{Detection examples from aerial images of TP (a) detections of open flowers and FP (b) from ripening fruits}
        \label{fig:Strawberry}
\end{figure}

Although TP detection proportion was low (37.1 percent) for YoloV5 ,overall, 97 percent of flowers were accurately detected (n=1218 of 1243 in dataset). However, the FP detections (n=2042,62.2 percent) far out numbered the TP leading to a lower proportion. 33.3\% (n=680) of FP detections were accounted for ripening or developing fruits \ref{fig:strawberry}b. A similar result was attained with Faster R-CNN ( 17.6 \%).YoloV5 showed overall best performance for our system. As our algorithm was not trained with strawberry fruits, and they exhibited similar spectral properties as the flowers, this lead to increased FP detections (see for instance Fig.~\ref{fig:Strawberry}). \cite{chen2019strawberry} conducted a similar flower detection study on strawberries using a Phantom 4 Pro to capture aerial images in RGB. Table \ref{table:compareresutls} compares their results with ours. \cite{chen2019strawberry} used Faster R-CNN with Resnet50 trained on Imagenet at a flight altitude of 3m. 

Our system has a higher mAP with Faster R-CNN as compared with their study, however, overall detection was low. Our YoloV5 algorithm vastly outperformed both Faster R-CNN algorithms in terms of training time (0.3 vs 4.5-5.5 hours) and image processing (0.0083 vs 1.86-8.872 FPS). mAP was also significantly Higher for both YoloV5 ad Faster R-CNN on our monochrome images\cite{chen2019strawberry} RGB images(mAp=0.951 vs mAP= 0.772).

\section{Discussion}

Farmers, like the ones from our test field, re-plant yearly for best harvest results. A tool for flower detection would need to be robust to changing varieties and cultivars and be able to provide consistent results.The test field was randomly planted with both 'Seascape' and 'Albion' cultivars. Although 'Seascape' was included in our initial training dataset 'Albion' was not. Our algorithm was able to  detect novel cultivar flowers as strawberry flowers. This indicates that a limited database of strawberry cultivar flowers could be sufficient for developing remote sensing tools and would only require limited updating to accommodate new varieties to the market.  


Schaefer, McGraw and Catoni \cite{schaefer2008birds} explored the effects of fruit color variation as signals of dietary reward. They found that fruits rich in anthocyanins, a plant antioxidant, are black or UV- reflecting and are in higher concentrations in ripe fruits. In strawberry fruits, anthocyanin concentration increases with fruit maturity \cite{da2007anthocyanin, song2015quantitative}. As our camera perceives the UV but not Red spectral range ripening strawberries appear bright white in our frames. When at similar size and circularity these fruits are mislabeled as flowers by our algorithm.


Remaining FP detections can be attributed to solar reflection on leaves, runners and flower stocks. Theses are expamples of the complexity of our field setting. Including null images of background foliage in initial training would also increase overall algorithm robustness. However, no FP were attributed to weed species which were visible within and between crop rows. These included: Fleabane, lambsquarter, crabgrass, purslane, and cow vetch. Some of these plants were in flower however the algorithm could distinguish between these and strawberry blossoms. This aspect may be explored in future research. A significant limitation to this research was the aerial resolution of the flowers. 

\section{Conclusion}
In this paper we've described and discussed the development of a cost- effective, light weight, UV-sensitive camera, used with a promising aerial platform. We performed extensive field experimentation to gather high quality imagery data and demonstrated its usability feeding it to two state-of-the-art object detection algorithms. The improved results over previous similar experiments show that our system is highly scalable, as the cost of the system is low, and that the UV spectrum can provide valuable information about the crops' flowers.

The development of this sensor and the choice of the aerial platform opens up the opportunities for a lot of potential work in the future. The development of a flower detector is the first step towards the bigger goals, such as estimating crop yields and creating a functional, biologically inspired robotic pollinator. Currently, using ROS onboard of Spiri Mu drone, the GPS coordinates are being recorded as well as rostopics. In the future, these GPS coordinates can be used in conjunction with the images to create a global map of the field and to direct harvest efforts. Additionally, developing automatic row cropping methodologies would help in getting better results with the object detection algorithms. Furtur iterations of this study will also include strawberry fruits at various developmental stages in the initial training process. Our sensor design shows a linear relationship between percent reflectance and pixel value between 300 and 650 nm opening up many possibilities for it's use even beyond agriculture. Although in this study we designed a filter which allowed 300-650 nm, the sensor can easily be modified into a UV-only camera with the addition of another filter such as the Xnite330 from Maxmax. Our design can be ordered to specification; and, the sensor transmission spectra is provided upon request. Lastly, our monochrome sensor design does not allow for the use of contrast between chroma channels as RGB sensors or animal eyes do. Furtur sensor iterations should explore removing only the red color of the Bayer filter. This would retain the green and blue channels and replace the red channel with UV using our existing camera body design. Computer analysis for 3 channel images could be carried out routinely as the proportion and placement of the three channels on the sensor diode are unchanged. Thus, a number of enhancements to this platform can improve the overall performance of the system and can help us make significant contributions to accelerate the research in this area. Agriculture has, and always will be, a sector for innovation. We openly make our datasets and trained algorithms available in hopes of furthering scientific pursuits.

\section*{Acknowledgment}
We must thank Spiri Robotics for lending us a platform to conduct our field deployment. Corentin Boucher and Ryan Brown were also key to the success of the prototype adaptation and the field deployment. We also thank Pr. Marcel Babin and his team at Takuvik in University of Laval, Canada for lending us their spectrophotometer to charaterize our sensor.

\bibliographystyle{IEEEtran}
\bibliography{main}

\begin{thebibliography}{10}
\providecommand{\url}[1]{#1}
\csname url@samestyle\endcsname
\providecommand{\newblock}{\relax}
\providecommand{\bibinfo}[2]{#2}
\providecommand{\BIBentrySTDinterwordspacing}{\spaceskip=0pt\relax}
\providecommand{\BIBentryALTinterwordstretchfactor}{4}
\providecommand{\BIBentryALTinterwordspacing}{\spaceskip=\fontdimen2\font plus
\BIBentryALTinterwordstretchfactor\fontdimen3\font minus
  \fontdimen4\font\relax}
\providecommand{\BIBforeignlanguage}[2]{{%
\expandafter\ifx\csname l@#1\endcsname\relax
\typeout{** WARNING: IEEEtran.bst: No hyphenation pattern has been}%
\typeout{** loaded for the language `#1'. Using the pattern for}%
\typeout{** the default language instead.}%
\else
\language=\csname l@#1\endcsname
\fi
#2}}
\providecommand{\BIBdecl}{\relax}
\BIBdecl

\bibitem{tilman2011global}
D.~Tilman, C.~Balzer, J.~Hill, and B.~L. Befort, ``Global food demand and the
  sustainable intensification of agriculture,'' \emph{Proceedings of the
  national academy of sciences}, vol. 108, no.~50, pp. 20\,260--20\,264, 2011.

\bibitem{tsouros2019review}
D.~C. Tsouros, S.~Bibi, and P.~G. Sarigiannidis, ``A review on uav-based
  applications for precision agriculture,'' \emph{Information}, vol.~10,
  no.~11, p. 349, 2019.

\bibitem{shakhatreh2019unmanned}
H.~Shakhatreh, A.~H. Sawalmeh, A.~Al-Fuqaha, Z.~Dou, E.~Almaita, I.~Khalil,
  N.~S. Othman, A.~Khreishah, and M.~Guizani, ``Unmanned aerial vehicles
  (uavs): A survey on civil applications and key research challenges,''
  \emph{Ieee Access}, vol.~7, pp. 48\,572--48\,634, 2019.

\bibitem{aurellunited}
D.~Aurell, S.~Bruckner, M.~Wilson, N.~Steinhauer, and G.~Williams, ``United
  states honey bee colony losses 2021-2022: Preliminary results from the bee
  informed partnership embargoed until thursday, july 28th, 2022, 12.00 pm noon
  est.''

\bibitem{goldstein2018green}
H.~Goldstein, ``The green promise of vertical farms [blueprints for a
  miracle],'' \emph{IEEE Spectrum}, vol.~55, no.~6, pp. 50--55, 2018.

\bibitem{kattenborn2021review}
T.~Kattenborn, J.~Leitloff, F.~Schiefer, and S.~Hinz, ``Review on convolutional
  neural networks (cnn) in vegetation remote sensing,'' \emph{ISPRS Journal of
  Photogrammetry and Remote Sensing}, vol. 173, pp. 24--49, 2021.

\bibitem{briscoe2001evolution}
A.~D. Briscoe and L.~Chittka, ``The evolution of color vision in insects,''
  \emph{Annual review of entomology}, vol.~46, no.~1, pp. 471--510, 2001.

\bibitem{virlet2016field}
N.~Virlet, K.~Sabermanesh, P.~Sadeghi-Tehran, and M.~J. Hawkesford, ``Field
  scanalyzer: An automated robotic field phenotyping platform for detailed crop
  monitoring,'' \emph{Functional Plant Biology}, vol.~44, no.~1, pp. 143--153,
  2016.

\bibitem{deery2014proximal}
D.~Deery, J.~Jimenez-Berni, H.~Jones, X.~Sirault, and R.~Furbank, ``Proximal
  remote sensing buggies and potential applications for field-based
  phenotyping,'' \emph{Agronomy}, vol.~4, no.~3, pp. 349--379, 2014.

\bibitem{williams2020improvements}
H.~Williams, C.~Ting, M.~Nejati, M.~H. Jones, N.~Penhall, J.~Lim, M.~Seabright,
  J.~Bell, H.~S. Ahn, A.~Scarfe \emph{et~al.}, ``Improvements to and
  large-scale evaluation of a robotic kiwifruit harvester,'' \emph{Journal of
  Field Robotics}, vol.~37, no.~2, pp. 187--201, 2020.

\bibitem{mulla2013twenty}
D.~J. Mulla, ``Twenty five years of remote sensing in precision agriculture:
  Key advances and remaining knowledge gaps,'' \emph{Biosystems engineering},
  vol. 114, no.~4, pp. 358--371, 2013.

\bibitem{kim2019unmanned}
J.~Kim, S.~Kim, C.~Ju, and H.~I. Son, ``Unmanned aerial vehicles in
  agriculture: A review of perspective of platform, control, and
  applications,'' \emph{Ieee Access}, vol.~7, pp. 105\,100--105\,115, 2019.

\bibitem{sankaran2015low}
S.~Sankaran, L.~R. Khot, C.~Z. Espinoza, S.~Jarolmasjed, V.~R. Sathuvalli,
  G.~J. Vandemark, P.~N. Miklas, A.~H. Carter, M.~O. Pumphrey, N.~R. Knowles
  \emph{et~al.}, ``Low-altitude, high-resolution aerial imaging systems for row
  and field crop phenotyping: A review,'' \emph{European Journal of Agronomy},
  vol.~70, pp. 112--123, 2015.

\bibitem{williams2020autonomous}
H.~Williams, M.~Nejati, S.~Hussein, N.~Penhall, J.~Y. Lim, M.~H. Jones,
  J.~Bell, H.~S. Ahn, S.~Bradley, P.~Schaare \emph{et~al.}, ``Autonomous
  pollination of individual kiwifruit flowers: Toward a robotic kiwifruit
  pollinator,'' \emph{Journal of Field Robotics}, vol.~37, no.~2, pp. 246--262,
  2020.

\bibitem{li2022identification}
K.~Li, L.~Zhai, H.~Pan, Y.~Shi, X.~Ding, and Y.~Cui, ``Identification of the
  operating position and orientation of a robotic kiwifruit pollinator,''
  \emph{Biosystems Engineering}, vol. 222, pp. 29--44, 2022.

\bibitem{le2020low}
T.~D. Le, V.~R. Ponnambalam, J.~G. Gjevestad, and P.~J. From, ``A low-cost and
  efficient autonomous row-following robot for food production in
  polytunnels,'' \emph{Journal of Field Robotics}, vol.~37, no.~2, pp.
  309--321, 2020.

\bibitem{ko2014autonomous}
M.~H. Ko, B.-S. Ryuh, K.~C. Kim, A.~Suprem, and N.~P. Mahalik, ``Autonomous
  greenhouse mobile robot driving strategies from system integration
  perspective: Review and application,'' \emph{IEEE/ASME Transactions On
  Mechatronics}, vol.~20, no.~4, pp. 1705--1716, 2014.

\bibitem{hong2012automatic}
S.-W. Hong and L.~Choi, ``Automatic recognition of flowers through color and
  edge based contour detection,'' in \emph{2012 3rd International conference on
  image processing theory, tools and applications (IPTA)}.\hskip 1em plus 0.5em
  minus 0.4em\relax IEEE, 2012, pp. 141--146.

\bibitem{senthilnath2016detection}
J.~Senthilnath, A.~Dokania, M.~Kandukuri, K.~Ramesh, G.~Anand, and S.~Omkar,
  ``Detection of tomatoes using spectral-spatial methods in remotely sensed rgb
  images captured by uav,'' \emph{Biosystems engineering}, vol. 146, pp.
  16--32, 2016.

\bibitem{antolinez2022identification}
A.~Antol{\'\i}nez~Garc{\'\i}a and J.~W. C{\'a}ceres~Campana, ``Identification
  of pathogens in corn using near-infrared uav imagery and deep learning,''
  \emph{Precision Agriculture}, pp. 1--24, 2022.

\bibitem{zhou2021assessment}
Z.~Zhou, Y.~Majeed, G.~D. Naranjo, and E.~M. Gambacorta, ``Assessment for crop
  water stress with infrared thermal imagery in precision agriculture: A review
  and future prospects for deep learning applications,'' \emph{Computers and
  Electronics in Agriculture}, vol. 182, p. 106019, 2021.

\bibitem{abdulridha2020detecting}
J.~Abdulridha, Y.~Ampatzidis, P.~Roberts, and S.~C. Kakarla, ``Detecting
  powdery mildew disease in squash at different stages using uav-based
  hyperspectral imaging and artificial intelligence,'' \emph{Biosystems
  Engineering}, vol. 197, pp. 135--148, 2020.

\bibitem{gomez2016field}
D.~G{\'o}mez-Cand{\'o}n, N.~Virlet, S.~Labb{\'e}, A.~Jolivot, and J.-L.
  Regnard, ``Field phenotyping of water stress at tree scale by uav-sensed
  imagery: new insights for thermal acquisition and calibration,''
  \emph{Precision agriculture}, vol.~17, no.~6, pp. 786--800, 2016.

\bibitem{stumph2019detecting}
B.~Stumph, M.~H. Virto, H.~Medeiros, A.~Tabb, S.~Wolford, K.~Rice, and
  T.~Leskey, ``Detecting invasive insects with unmanned aerial vehicles,'' in
  \emph{2019 International Conference on Robotics and Automation (ICRA)}.\hskip
  1em plus 0.5em minus 0.4em\relax IEEE, 2019, pp. 648--654.

\bibitem{dyer2015seeing}
A.~G. Dyer, J.~E. Garcia, M.~Shrestha, and K.~Lunau, ``Seeing in colour: a
  hundred years of studies on bee vision since the work of the nobel laureate
  karl von frisch,'' \emph{Proceedings of the Royal Society of Victoria}, vol.
  127, no.~1, pp. 66--72, 2015.

\bibitem{coliban2020linear}
R.-M. Coliban, M.~Marinca{\c{s}}, C.~Hatfaludi, and M.~Ivanovici, ``Linear and
  non-linear models for remotely-sensed hyperspectral image visualization,''
  \emph{Remote Sensing}, vol.~12, no.~15, p. 2479, 2020.

\bibitem{arnold2008fred}
S.~Arnold, V.~Savolainen, and L.~Chittka, ``Fred: the floral reflectance
  spectra database,'' \emph{Nature Precedings}, pp. 1--1, 2008.

\bibitem{stuart2019hyperspectral}
M.~B. Stuart, A.~J. McGonigle, and J.~R. Willmott, ``Hyperspectral imaging in
  environmental monitoring: a review of recent developments and technological
  advances in compact field deployable systems,'' \emph{Sensors}, vol.~19,
  no.~14, p. 3071, 2019.

\bibitem{canadastrawberry}
\BIBentryALTinterwordspacing
Statistical overview of the canadian fruit industry 2020. [Online]. Available:
  \url{https://agriculture.canada.ca/en/canadas-agriculture-sectors/horticulture/horticulture-sector-reports/statistical-overview-canadian-fruit-industry-2020}
\BIBentrySTDinterwordspacing

\bibitem{James123}
\BIBentryALTinterwordspacing
R.~James and T.~L. Pitts-Singer, \emph{{Bee Pollination in Agricultural
  Ecosystems}}.\hskip 1em plus 0.5em minus 0.4em\relax Oxford University Press,
  07 2008. [Online]. Available:
  \url{https://doi.org/10.1093/acprof:oso/9780195316957.001.0001}
\BIBentrySTDinterwordspacing

\bibitem{palum2001image}
R.~Palum, ``Image sampling with the bayer color filter array,'' in \emph{PICS},
  2001, pp. 239--245.

\bibitem{vanbrabant2020pear}
Y.~Vanbrabant, S.~Delalieux, L.~Tits, K.~Pauly, J.~Vandermaesen, and B.~Somers,
  ``Pear flower cluster quantification using rgb drone imagery,''
  \emph{Agronomy}, vol.~10, no.~3, p. 407, 2020.

\bibitem{chen2019strawberry}
Y.~Chen, W.~S. Lee, H.~Gan, N.~Peres, C.~Fraisse, Y.~Zhang, and Y.~He,
  ``Strawberry yield prediction based on a deep neural network using
  high-resolution aerial orthoimages,'' \emph{Remote Sensing}, vol.~11, no.~13,
  p. 1584, 2019.

\bibitem{hunt2018good}
E.~R. Hunt~Jr and C.~S. Daughtry, ``What good are unmanned aircraft systems for
  agricultural remote sensing and precision agriculture?'' \emph{International
  journal of remote sensing}, vol.~39, no. 15-16, pp. 5345--5376, 2018.

\bibitem{maxmaxcam}
\BIBentryALTinterwordspacing
Maxmax monochrome camera module. [Online]. Available:
  \url{https://maxmax.com/shopper/product/15991-xniteusb8m-m-usb-2-0-8-megapixel-monchrome-camera-module/category\_pathway-9532}
\BIBentrySTDinterwordspacing

\bibitem{goprocamera}
\BIBentryALTinterwordspacing
Gopro hero9 black action camera. [Online]. Available:
  \url{https://www.juzaphoto.com/recensione.php? l=en\&t=gopro\_hero9\_black}
\BIBentrySTDinterwordspacing

\bibitem{graph1}
\BIBentryALTinterwordspacing
Schott n-wg280 colorless longpass filter. [Online]. Available:
  \url{https://schott.com/shop/advanced-optics/en/Matt-Filter-Plates/N-WG2N-WG28080/c/glass-N-WG280}
\BIBentrySTDinterwordspacing

\bibitem{sensor1}
\BIBentryALTinterwordspacing
Monochrome vs color sensors. [Online]. Available:
  \url{https://www.opto-e.com/basics/monochrome-vs-color-sensors}
\BIBentrySTDinterwordspacing

\bibitem{dyer2004calibrated}
A.~G. Dyer, L.~Muir, and W.~Muntz, ``A calibrated gray scale for forensic
  ultraviolet photography,'' \emph{Journal of Forensic Science}, vol.~49,
  no.~5, pp. JFS2\,003\,410--3, 2004.

\bibitem{zheng2021remote}
C.~Zheng, A.~Abd-Elrahman, and V.~Whitaker, ``Remote sensing and machine
  learning in crop phenotyping and management, with an emphasis on applications
  in strawberry farming,'' \emph{Remote Sensing}, vol.~13, no.~3, p. 531, 2021.

\bibitem{lin2018detection}
P.~Lin and Y.~Chen, ``Detection of strawberry flowers in outdoor field by deep
  neural network,'' in \emph{2018 IEEE 3rd International Conference on Image,
  Vision and Computing (ICIVC)}.\hskip 1em plus 0.5em minus 0.4em\relax IEEE,
  2018, pp. 482--486.

\bibitem{zhou2020strawberry}
C.~Zhou, W.~S. Lee, and R.~Lin, ``Strawberry flower detection using
  fluorescence imaging,'' in \emph{2020 ASABE Annual International Virtual
  Meeting}.\hskip 1em plus 0.5em minus 0.4em\relax American Society of
  Agricultural and Biological Engineers, 2020, p.~1.

\bibitem{mahendrakar2022performance}
T.~Mahendrakar, A.~Ekblad, N.~Fischer, R.~White, M.~Wilde, B.~Kish, and
  I.~Silver, ``Performance study of yolov5 and faster r-cnn for autonomous
  navigation around non-cooperative targets,'' in \emph{2022 IEEE Aerospace
  Conference (AERO)}.\hskip 1em plus 0.5em minus 0.4em\relax IEEE, 2022, pp.
  1--12.

\bibitem{immaneni2022real}
A.~Immaneni and Y.~K. Chang, ``Real-time counting of strawberry using
  cost-effective embedded gpu and yolov4-tiny,'' in \emph{2022 ASABE Annual
  International Meeting}.\hskip 1em plus 0.5em minus 0.4em\relax American
  Society of Agricultural and Biological Engineers, 2022, p.~1.

\bibitem{bochkovskiy2020yolov4}
A.~Bochkovskiy, C.-Y. Wang, and H.-Y.~M. Liao, ``Yolov4: Optimal speed and
  accuracy of object detection,'' \emph{arXiv preprint arXiv:2004.10934}, 2020.

\bibitem{wang2022study}
C.~Wang, Y.~Wang, S.~Liu, G.~Lin, P.~He, Z.~Zhang, and Y.~Zhou, ``Study on pear
  flowers detection performance of yolo-pefl model trained with synthetic
  target images,'' \emph{Frontiers in Plant Science}, vol.~13, 2022.

\bibitem{strawberryvarieties}
\BIBentryALTinterwordspacing
Strawberry varieties for quebec. [Online]. Available:
  \url{https://strawberryplants.org/recommended-strawberry-varieties-for-canada/\#QC}
\BIBentrySTDinterwordspacing

\bibitem{roboflowyolo}
\BIBentryALTinterwordspacing
Roboflow. [Online]. Available: \url{https://roboflow.com/}
\BIBentrySTDinterwordspacing

\bibitem{glenn_jocher_2020_4154370}
\BIBentryALTinterwordspacing
G.~Jocher, A.~Stoken, J.~Borovec, NanoCode012, ChristopherSTAN, L.~Changyu,
  Laughing, tkianai, A.~Hogan, lorenzomammana, yxNONG, AlexWang1900,
  L.~Diaconu, Marc, wanghaoyang0106, ml5ah, Doug, F.~Ingham, Frederik, Guilhen,
  Hatovix, J.~Poznanski, J.~Fang, L.~Yu, changyu98, M.~Wang, N.~Gupta,
  O.~Akhtar, PetrDvoracek, and P.~Rai, ``{ultralytics/yolov5: v3.1 - Bug Fixes
  and Performance Improvements},'' Oct. 2020. [Online]. Available:
  \url{https://doi.org/10.5281/zenodo.4154370}
\BIBentrySTDinterwordspacing

\bibitem{ren2015faster}
S.~Ren, K.~He, R.~Girshick, and J.~Sun, ``Faster r-cnn: Towards real-time
  object detection with region proposal networks,'' \emph{Advances in neural
  information processing systems}, vol.~28, 2015.

\bibitem{lin2014microsoft}
T.-Y. Lin, M.~Maire, S.~Belongie, J.~Hays, P.~Perona, D.~Ramanan,
  P.~Doll{\'a}r, and C.~L. Zitnick, ``Microsoft coco: Common objects in
  context,'' in \emph{European conference on computer vision}.\hskip 1em plus
  0.5em minus 0.4em\relax Springer, 2014, pp. 740--755.

\bibitem{spiriweb}
\BIBentryALTinterwordspacing
Spiri mu. [Online]. Available:
  \url{https://spirirobotics.com/products/spiri-mu/}
\BIBentrySTDinterwordspacing

\bibitem{Bisong2019}
\BIBentryALTinterwordspacing
E.~Bisong, \emph{Google Colaboratory}.\hskip 1em plus 0.5em minus 0.4em\relax
  Berkeley, CA: Apress, 2019, pp. 59--64. [Online]. Available:
  \url{https://doi.org/10.1007/978-1-4842-4470-8\_7}
\BIBentrySTDinterwordspacing

\bibitem{mavric3specs}
\BIBentryALTinterwordspacing
Dji mavric 3 specs. [Online]. Available:
  \url{https://www.dji.com/ca/mavic-3/specs}
\BIBentrySTDinterwordspacing

\bibitem{matrice300specs}
\BIBentryALTinterwordspacing
Dji matrice 300 specifications and zenmuse h20 hybrid thermal camera. [Online].
  Available:
  \url{https://dronedj.com/2020/01/21/dji-matrice-300-specifications-zenmuse-h20-hybrid-camera/}
\BIBentrySTDinterwordspacing

\bibitem{phantom4specs}
\BIBentryALTinterwordspacing
Dji phantom 4 pro. [Online]. Available:
  \url{https://www.dji.com/ca/phantom-4-pro/info}
\BIBentrySTDinterwordspacing

\bibitem{schaefer2008birds}
H.~M. Schaefer, K.~McGraw, and C.~Catoni, ``Birds use fruit colour as honest
  signal of dietary antioxidant rewards,'' \emph{Functional Ecology}, vol.~22,
  no.~2, pp. 303--310, 2008.

\bibitem{da2007anthocyanin}
F.~L. da~Silva, M.~T. Escribano-Bail{\'o}n, J.~J.~P. Alonso, J.~C.
  Rivas-Gonzalo, and C.~Santos-Buelga, ``Anthocyanin pigments in strawberry,''
  \emph{LWT-Food Science and Technology}, vol.~40, no.~2, pp. 374--382, 2007.

\bibitem{song2015quantitative}
J.~Song, L.~Du, L.~Li, W.~Kalt, L.~C. Palmer, S.~Fillmore, Y.~Zhang, Z.~Zhang,
  and X.~Li, ``Quantitative changes in proteins responsible for flavonoid and
  anthocyanin biosynthesis in strawberry fruit at different ripening stages: a
  targeted quantitative proteomic investigation employing multiple reaction
  monitoring,'' \emph{Journal of proteomics}, vol. 122, pp. 1--10, 2015.

\end{thebibliography}

\end{document}